\documentclass{article}
\usepackage{url}
\usepackage{color}
\usepackage{layouts}
\usepackage{graphicx}
\usepackage{amsmath}
\usepackage{amsfonts}
\usepackage{inputenc}
\usepackage{mathtools}
\usepackage{multirow}
\usepackage{times}
\usepackage{natbib}
\usepackage{algorithm}
\usepackage{algorithmic}
\usepackage{hyperref}

\usepackage{booktabs}
\usepackage{caption}

\usepackage[accepted]{icml2017}

\definecolor{light-gray}{gray}{0.4}
\definecolor{dark-green}{rgb}{0,0.4,0}

\icmltitlerunning{Device Placement Optimization with Reinforcement Learning}

\begin{document}

\twocolumn[
\icmltitle{Device Placement Optimization with Reinforcement Learning}

\icmlsetsymbol{equal}{*}

\begin{icmlauthorlist}
\icmlauthor{Azalia Mirhoseini}{equal,Google Brain,Residency}
\icmlauthor{Hieu Pham}{equal,Google Brain,Residency}
\icmlauthor{Quoc V. Le}{Google Brain}
\icmlauthor{Benoit Steiner}{Google Brain}
\icmlauthor{Rasmus Larsen}{Google Brain}
\icmlauthor{Yuefeng Zhou}{Google Brain}
\icmlauthor{Naveen Kumar}{Google}
\icmlauthor{Mohammad Norouzi}{Google Brain}
\icmlauthor{Samy Bengio}{Google Brain}
\icmlauthor{Jeff Dean}{Google Brain}
\end{icmlauthorlist}

\icmlaffiliation{Google Brain}{Google Brain}
\icmlaffiliation{Google}{Google}
\icmlaffiliation{Residency}{Members of the Google Brain Residency Program
      (\url{g.co/brainresidency})}

\icmlcorrespondingauthor{Azalia Mirhoseini}{azalia@google.com}
\icmlcorrespondingauthor{Hieu Pham}{hyhieu@google.com}

\icmlkeywords{Reinforcement Learning, Device Placement}

\vskip 0.3in
]

\printAffiliationsAndNotice{\icmlEqualContribution}

\begin{abstract}
The past few years have witnessed a growth in size and computational
requirements for training and inference with neural networks. Currently, a
common approach to address these requirements is to use a heterogeneous
distributed environment with a mixture of hardware devices such as CPUs and
GPUs. Importantly, the decision of placing parts of the neural models on
devices is often made by human experts based on simple heuristics and
intuitions. In this paper, we propose a method which learns to optimize device
placement for TensorFlow computational graphs.  Key to our method is the use of
a sequence-to-sequence model to predict which subsets of operations in a
TensorFlow graph should run on which of the available devices. The execution
time of the predicted placements is then used as the reward signal to optimize
the parameters of the sequence-to-sequence model. Our main result is that on
Inception-V3 for ImageNet classification, and on RNN LSTM, for language
modeling and neural machine translation, our model finds non-trivial device
placements that outperform hand-crafted heuristics and traditional algorithmic
methods.

\end{abstract}

\section{\label{sec:intro}Introduction}
Over the past few years, neural networks have proven to be a general and
effective tool for many practical problems, such as image
classification~\cite{krizhevsky2012imagenet,szegedy2015going,he2016deep},
speech
recognition~\cite{hinton2012deep,graves2014towards,hannun2014deep,chan2015listen},
machine
translation~\cite{sutskever2014sequence,cho2014learning,bahdanau15attention,gnmt}
and speech synthesis~\cite{oord2016wavenet,arik2017deep,wang2017}.  Together
with their success is the growth in size and computational requirements of
training and inference.  Currently, a typical approach to address these
requirements is to use a heterogeneous distributed environment with a mixture
of many CPUs and GPUs. In this environment, it is a common practice for a
machine learning practitioner to specify the device placement for certain
operations in the neural network.  For example, in a neural translation
network, each layer, including all LSTM layers, the attention layer, and the
softmax layer, is computed by a GPU~\cite{sutskever2014sequence,gnmt}.

Although such decisions can be made by machine learning practitioners, they can
be challenging, especially when the network has many branches~\cite{inception},
or when the minibatches get larger. Existing algorithmic
solvers~\cite{scotch,karypis1995metis}, on the other hand, are not flexible
enough to work with a dynamic environment with many interferences.

\begin{figure}[!h]
   \centering
   \includegraphics[width=0.4\textwidth]{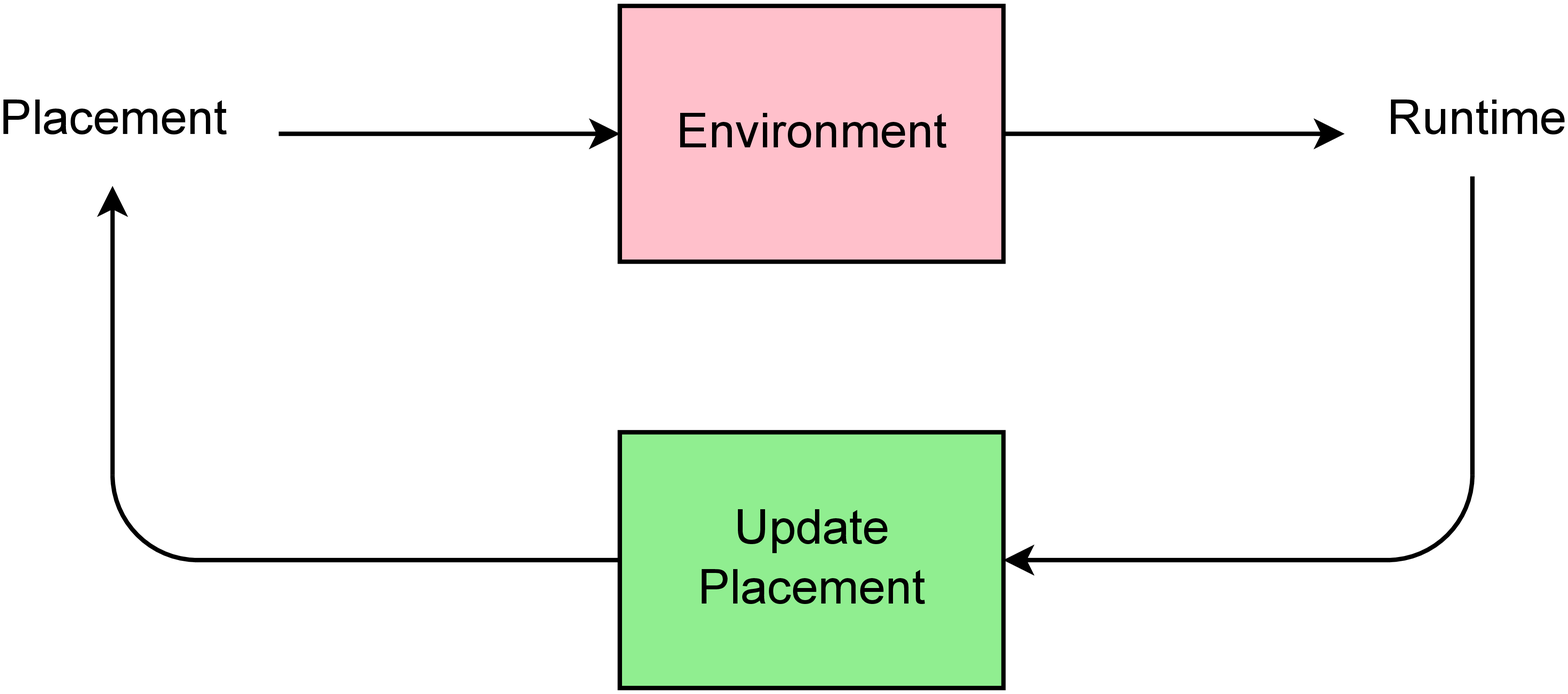}
   \caption{An overview of the RL based device placement model.}
    \label{fig:overview}
\end{figure}
In this paper, we propose a method which learns to optimize device placement
for training and inference with neural networks.  The method, illustrated in
Figure~\ref{fig:overview}, takes into account information of the environment by
performing series of experiments to understand which parts of the model should
be placed on which device, and how to arrange the computations so that the
communication is optimized. Key to our method is the use of a
sequence-to-sequence model to read input information about the operations as
well as the dependencies between them, and then propose a placement for each
operation.  Each proposal is executed in the hardware environment to measure
the execution time.  The execution time is then used as a reward signal to
train the recurrent model so that it gives better proposals over time.

Our main result is that our method finds non-trivial placements on multiple
devices for Inception-V3~\cite{inception}, Recurrent Neural Language
Model~\cite{zaremba14,lm1b} and Neural Machine
Translation~\cite{sutskever2014sequence,gnmt}. Single-step measurements show
that Scotch~\citep{scotch} yields disappointing results on all three
benchmarks, suggesting that their graph-based heuristics are not flexible
enough for them. Our method can find non-trivial placements that are up to
$3.5$ times faster. When applied to train the three models in real time, the
placements found by our method are up to $20\%$ faster than human experts'
placements.

\section{\label{sec:related}Related Work}
Our work is closely related to the idea of using neural networks and
reinforcement learning for combinatorial
optimization~\cite{vinyals2015pointer,bello2016}. The space of possible
placements for a computational graph is discrete, and we model the placements
using a sequence-to-sequence approach, trained with policy gradients. However,
experiments in early work were only concerned with toy datasets, whereas this
work applies the framework to a large-scale practical application with noisy
rewards.

Reinforcement learning has also been applied to optimize system performance.
For example,~\citet{rl_alloc16} propose to train a resource management
algorithm with policy gradients. However, they optimize the expected value of a
hand-crafted objective function based on the reward, unlike this work, where we
optimize directly for the running time of the configurations, hence relieving
the need to design intermediate cost models.

Graph partitioning is an intensively studied subject in computer science. Early
work such as
\citet{kernighan1970efficient,kirkpatrick1983optimization,fiduccia1988linear,johnson1989optimization}
employ several iterative refinement procedures that start from a partition and
continue to explore similar partitions to improve. Alternative methods such as
\citet{hagen1992new,karypis1995metis} perform spectral analyses on matrix
representations of graphs to partition them. Despite their extensive
literature, graph partitioning algorithms remain heuristics for computational
graphs. The reason is that in order to apply these algorithms, one has to
construct cost models for the graphs of concern. Since such models are
expensive to even estimate and in virtually all cases, are not accurate, graph
partitioning algorithms applied on them can lead to unsatisfying results, as we
show in Section~\ref{sec:exp} of this paper.

A well-known graph partitioning algorithm with an open source software library
is the Scotch optimizer~\cite{scotch}, which we use as a baseline in our
experiments. The Scotch mapper attempts to balance the computational load of a
collection of tasks among a set of connected processing nodes, while reducing
the cost of communication by keeping intensively communicating tasks on nearby
nodes. Scotch relies on a collection of graph partitioning techniques such as
k-way Fiduccia-Mattheyses~\cite{fiduccia1988linear}, multilevel
method~\cite{multilevel1,multilevel2,multilevel3}, band
method~\cite{bandmethod}, diffusion method~\cite{diffusionmethod}, and dual
recursive bipartitioning mapping~\cite{recursivebipartition}).

Scotch models the problem with $2$ graphs. The first graph is called the target
architecture graph, whose vertices represent hardware resources such as CPUs or
GPUs and whose edges represent the communication paths available between them,
such as a PCIe bus or a network link. The second graph is called the source
graph, which models the computation to be mapped onto the target architecture
graph.  In the case of TensorFlow~\citep{tensorflow}, the computations of
programs are modeled as a graph whose vertices represent operations, while the
graph edges represent the multidimensional data arrays (tensors) communicated
between them. Scotch users have to choose how and when given partitioning
should be applied to graphs.  However, in our experiment, we rely on the
software's default strategies implemented in Scotch, which have already been
extensively tuned.

\section{\label{sec:method}Method}

Consider a TensorFlow computational graph $\mathcal{G}$, which consists of $M$
operations $\{o_1, o_2, ..., o_M\}$, and a list of $D$ available devices. A
placement $\mathcal{P} = \{p_1, p_2, ..., p_M\}$ is an assignment of 
an operation $o_i \in \mathcal{G}$ to a device $p_i$, where $p_i \in \{1, ..., D\}$. Let
$r(\mathcal{P})$ denote the time that it takes to perform a complete execution of
$\mathcal{G}$ under the placement $\mathcal{P}$. The goal of \emph{device
placement optimization} is to find $\mathcal{P}$ such that the execution time
$r(\mathcal{P})$ is minimized.


\subsection{\label{sec:reinforce}Training with Policy Gradients}
\begin{figure*}
   \centering
   \includegraphics[width=\textwidth]{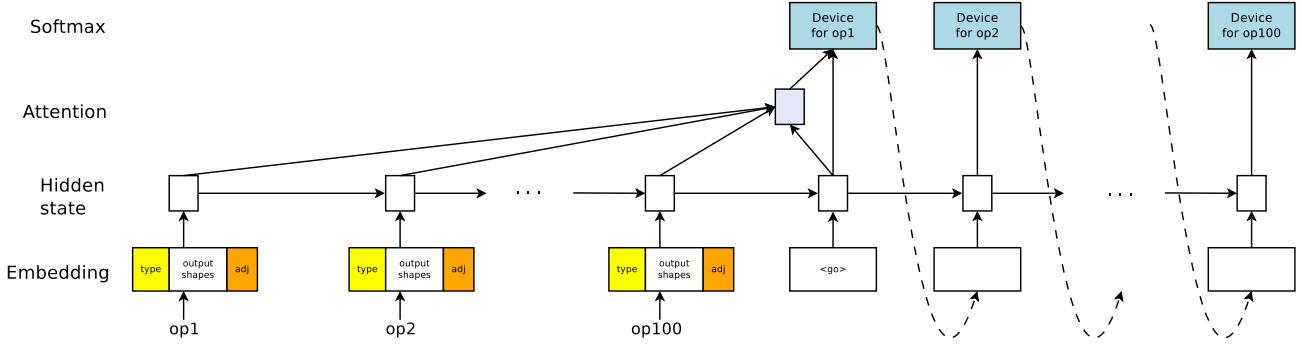}
      \caption{Device placement model architecture.}
         \label{fig:model}
\end{figure*}

While we seek to minimize the execution time $r(\mathcal{P})$, direct
optimization of $r(\mathcal{P})$ results in two major issues. First, in the
beginning of the training process, due to the bad placements sampled, the
measurements of $r(\mathcal{P})$ can be noisy, leading to inappropriate
learning signals. Second, as the RL model gradually converges, the placements
that are sampled become more similar to each other, leading to small
differences between the corresponding running times, which results in less
distinguishable training signals. We  
empirically find that the square root of running time, $R(\mathcal{P}) =
\sqrt{r(\mathcal{P})}$, makes the learning process more robust. Accordingly, we
propose to train a stochastic policy
$\pi(\mathcal{P} | \mathcal{G}; \theta)$ to minimize the objective
\begin{align}
  \label{eqn:j_theta}
  J(\theta) = \mathbf{E}_{\mathcal{P} \sim \pi(\mathcal{P} | \mathcal{G}; \theta)} \left[R\left(\mathcal{P}\right) | \mathcal{G}\right]
\end{align}

In our work, $\pi(\mathcal{P} | \mathcal{G}; \theta)$ is defined by an
attentional sequence-to-sequence model, which we will describe in
Section~\ref{sec:model}.  We learn the network parameters using
Adam~\citep{adam} optimizer based on policy gradients computed via the
REINFORCE equation~\citep{reinforce},
\begin{align}
  \label{eqn:j_theta_grad}
  \nabla_\theta J(\theta) = \mathbf{E}_{\mathcal{P} \sim \pi(\mathcal{P} | \mathcal{G}; \theta)}
    \left[ R\left(\mathcal{P}\right) \cdot \nabla_\theta \log{p \left(\mathcal{P} | \mathcal{G}; \theta \right)} \right]
\end{align}

We estimate $\nabla_\theta J(\theta)$ by drawing $K$ placement samples using
$\mathcal{P}_i \sim \pi(\cdot | \mathcal{G}; \theta)$. We reduce the variance
of policy gradients by using a baseline term $B$, leading to
\begin{align}
  \label{eqn:j_theta_grad_baseline}
  \nabla_\theta J(\theta) ~\approx \frac{1}{K} \sum_{i=1}^{K} \left( R\left(\mathcal{P}_i\right)-B\right) \cdot \nabla_\theta \log{p \left(\mathcal{P}_i | \mathcal{G}; \theta \right)}
\end{align}

We find that a simple moving average baseline $B$ works well in our
experiments.  In practice, on computational graphs with large memory
footprints, some placements can fail to execute, e.g., putting all of the
operations of a huge LSTM on a single GPU will exceed the device's memory
limit. For such cases, we set the square root of running time $R(\mathcal{P})$
to a large constant, which we call the failing signal. We specify the failing
signal manually depending on the input graph. We observe that throughout our
training process, some placements sporadically and unexpectedly fail, perhaps
due to factors such as the state of the machine (we train our model on a shared
cluster). This phenomenon is particularly undesirable towards the end of the
training process, since a large difference between $R(\mathcal{P}_i)$ and the
baseline $B$ leads to a large update of the parameters, which potentially
perturbs parameters $\theta$ out of a good minimum. We thus hard-code the
training process so that after $5,000$ steps, one performs a parameter update
with a sampled placement $\mathcal{P}$ only if the placement executes. In our
experiments, we also find that initializing the baseline $B$ with the failing
signal results in more exploration.

\subsection{\label{sec:model}Architecture Details}
We use a sequence-to-sequence model~\citep{sutskever2014sequence} with
LSTM~\citep{lstm97} and a content-based attention
mechanism~\citep{bahdanau15attention} to predict the placements.
Figure~\ref{fig:model} shows the overall architecture of our model, which can
be divided into two parts: encoder RNN and decoder RNN. 

The input to the encoder RNN is the sequence of operations of the input graph.
We embed the operations by concatenating their information. Specifically, for
each input graph $\mathcal{G}$, we first collect the types of its
operations. An operation's type describes the underlying computation, such as
{\tt MatMul} or {\tt conv2d}. For each type, we store a tunable embedding
vector. We then record the size of each operation's list of output tensors and
concatenate them into a fixed-size zero-padded list called the output shape. We
also take the one-hot encoding vector that represents the operations that are
direct inputs and outputs to each operation. Finally, the embedding of each
operation is the concatenation of its type, its output shape, and its
one-hot encoded adjacency information.

The decoder is an attentional LSTM~\citep{bahdanau15attention} with a fixed
number of time steps that is equal to the number of operations in a graph
$\mathcal{G}$. At each step, the decoder outputs the device for the operation
at the same encoder time step. Each device has its own tunable embedding,
which is then fed as input to the next decoder time step.

\subsection{Co-locating Operations}
A key challenge when applying our method to TensorFlow computational graphs is
that these graphs generally have thousands of operations (see
Table~\ref{tab:model_stats}).  Modeling such a large number of operations with
sequence-to-sequence models is difficult due to vanishing and exploding
gradient issues~\citep{pascanu2013} and large memory footprints. We propose to
reduce the number of objects to place on different devices by manually forcing
several operations to be located on the same device. In practice, this is
implemented by the {\tt colocate\_with} feature of TensorFlow.

We use several heuristics to create co-location groups. First, we rely on
TensorFlow's default co-location groups, such as co-locating each operation's
outputs with its gradients. We further apply a simple heuristic to merge more
operations into co-location groups. Specifically, if the output of an operation
$X$ is consumed {\em only} by another operation $Y$, then operations $X$ and
$Y$ are co-located. Many initialization operations in TensorFlow can be grouped
in this way. In our experiments, we apply this heuristic recursively, and after
each iteration, we treat the co-location groups as operations until there are
not any further groups that can be merged. For certain models, we apply
specific rules to construct co-location groups. For example, with ConvNets, we
can treat several convolutions and pooling layers as a co-location group, and
with RNN models, we treat each LSTM cell as a group.

\subsection{Distributed Training}
\begin{figure}[ht]
  \centering
  \includegraphics[width=0.45\textwidth]{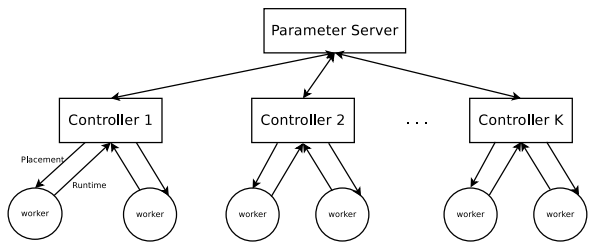}
  \caption{\label{fig:distributed}Distributed and asynchronous parameter update
  and reward evaluation.}
\end{figure}

We speed up the training process of our model using asynchronous distributed
training, as shown in Figure~\ref{fig:distributed}. Our framework consists of
several controllers, each of which execute the current policy defined
by the attentional sequence-to-sequence model as described in
Section~\ref{sec:model}. All of the controllers interact with a single shared
parameter server. We note that the parameter server holds only the controllers'
parameters, and not the input graph's parameters, because keeping the input
graph's parameters on the parameter server can potentially create a latency
bottleneck to transfer these parameters. Each controller in our framework
interacts with $K$ workers, where $K$ is the number of Monte Carlo samples in
Equation~\ref{eqn:j_theta_grad_baseline}.

The training process has two alternating phases. In the first phase, each
worker receives a signal that indicates that it should wait for placements from
its controller, while each controller receives a signal that indicates it
should sample $K$ placements. Each sampled placement comes with a probability.
Each controller then independently sends the placements to their workers, one
placement per worker, and sends a signal to indicate a phase change.

In the second phase, each worker  executes the placement it receives and
measures the running time. To reduce the variance in these measurements, each
placement is executed for $10$ steps and the average running time of the steps
but the first one is recorded. We observe that in TensorFlow, the first step
can take longer to execute compared to the following steps, and hence we treat
itss runing time as an outlier. Each controller waits for all of its workers to
finish executing their assigned placements and returning their running times.
When all of the running times are received, the controller uses the running
times to scale the corresponding gradients to asynchronously update the
controller parameters that reside in the parameter server.

In our experiments, we use up to $20$ controllers, each with either $4$ or $8$
workers. Under this setting, it takes between $12$ to $27$ hours to find the
best placement for the models in our experiments.  Using more workers per
controller yields more accurate estimates of the policy gradient as in
Equation~\ref{eqn:j_theta_grad_baseline}, but comes at the expense of possibly
having to put more workers in idle states. We also note that due to the
discrepancies between machines, it is more stable to let each controller have
its own baseline.

\section{\label{sec:exp}Experiments}
In the following experiments, we apply our proposed method to assign
computations to devices on three important neural networks in the deep learning
literature: Recurrent Neural Language Model (RNNLM)~\citep{zaremba14,lm1b},
Attentional Neural Machine Translation~\citep{bahdanau15attention}, and
Inception-V3~\citep{inception}. We compare the RL placements against strong
existing baselines described in Section~\ref{sec:baselines}.

\subsection{\label{sec:tasks}Experiment Setup}

\paragraph{Benchmarks.} We evaluate our approach on three established deep
learning models:
\begin{itemize}
  \item Recurrent Neural Network Language Model (RNNLM) with multiple LSTM
    layers~\citep{zaremba14,lm1b}. The grid structure of this model introduces
    tremendous potential for parallel executions because each LSTM cell can
    start as soon as its input and previous states are available.

  \item Neural Machine Translation with attention mechanism
    (NMT)~\citep{bahdanau15attention,gnmt}. While the architecture of this
    model is similar to that of RNNLM, its large number of hidden states due to
    the source and target sentences necessitates model parallelism.
    Both~\citet{sutskever2014sequence} and~\citet{gnmt} propose to place each
    LSTM layer, the attention layer, and the softmax layer, each on a separate
    device. While the authors observe significant improvements at training
    time, their choices are not optimal. In fact, we show in
    our experiments that a trained policy can find significantly better
    placements.

  \item Inception-V3~\citep{inception} is a widely-used architecture for
    image recognition and visual feature extraction~\citep{inception_app1,inception_app2}.
    The Inception network has
    multiple blocks.  Each block has several branches of convolutional and
    pooling layers, which are then concatenated to make the inputs for the next
    block. While these branches can be executed in parallel, the network's
    depth restricts such potential since the later blocks have to wait for the
    previous ones.
\end{itemize}

\paragraph{Model details.} For Inception-V3, each step is executed on a batch
of images, each of size $299 \times 299 \times 3$, which is the widely-used
setting for the ImageNet Challenge~\citep{szegedy2015going}. For RNNLM and NMT,
we use the model with $2$ LSTM layers, with sizes of $2048$ and $1024$,
respectively. We set the number of unrolling steps for RNNLM, as well as the
maximum length for the source and target sentences of NMT, to $40$. Each pass
on RNNLM and NMT consists of a minibatch of $64$ sequences.

\paragraph{Co-location groups.} We pre-process the TensorFlow computational
graphs of the three aforementioned models to manually create their co-location
groups. More precisely; for RNNLM and NMT, we treat each LSTM cell, each
embedding lookup, each attention step and each softmax prediction step as a
group; for Inception-V3, we treat each branch as a group.
Table~\ref{tab:model_stats} shows the grouping statistics of these models.
\begin{table}[h]
  \centering
  \begin{tabular}{l|ll}
    \toprule
    Model        & \#operations & \#groups \\
    \midrule[0.08em]
    RNNLM        & 8943         & 188      \\
    NMT          & 22097        & 280      \\
    Inception-V3 & 31180        & 83       \\
    \bottomrule
  \end{tabular}
  \caption{\label{tab:model_stats}Model statistics.}
\end{table}

\begin{table*}[htb]
  \centering
  \begin{tabular}{l|ll|llll|ll}
    \toprule
    Tasks        & Single-CPU   & Single-GPU                   & \#GPUs  & Scotch & MinCut & Expert & RL-based   & Speedup \\
    \midrule
    RNNLM        & \multirow{2}{*}{6.89}  & \multirow{2}{*}{\textbf{1.57}}  & 2       & 13.43  & 11.94  & 3.81  & \textbf{1.57} & 0.0\%          \\
    (batch $64$)                             &           &                     & 4       & 11.52  & 10.44  & 4.46  & \textbf{1.57} & 0.0\%          \\
    \midrule
    NMT          & \multirow{2}{*}{10.72}  & \multirow{2}{*}{OOM} & 2       & 14.19  & 11.54  & 4.99  & \textbf{4.04} & 23.5\%       \\
    (batch $64$) &          &                           & 4       & 11.23  & 11.78  & 4.73  & \textbf{3.92} & 20.6\%       \\
    \midrule[0.08em]
    Inception-V3 & \multirow{2}{*}{26.21} & \multirow{2}{*}{\textbf{4.60}}  & 2       & 25.24  & 22.88  & 11.22 & \textbf{4.60} & 0.0\%          \\
    (batch $32$) &                        &                        & 4       & 23.41  & 24.52  & 10.65 & \textbf{3.85} & 19.0\%       \\
    \bottomrule
  \end{tabular}
  \caption{\label{tab:results}Running times (in seconds) of placements found by
  RL-based method and the baselines (lower is better). For each model, the
  first row shows the results with 1 CPU and 2 GPUs; the second row shows the
  results with 1 CPU and 4 GPUs. Last column shows improvements in running time
  achieved by RL-based placement over fastest baseline. To reduce variance,
  running times less than $10$ seconds are measured $15$ times and the averages
  are recorded.  OOM is Out Of Memory.}
\end{table*}

\paragraph{Metrics.} We implement training operations for RNNLM and NMT using
Adam~\citep{adam}, and for Inception-V3 using RMSProp~\citep{Tieleman2012}.  We
evaluate a placement by the total time it takes to perform one forward pass,
one backward pass and one parameter update. To reduce measurement variance, we
average the running times over several trials.  Additionally, we train each
model from scratch using the placements found by our method and compare the
training time to that of the strongest baseline placement.

\paragraph{Devices.} In our experiments, the available devices are 1 Intel
Haswell 2300 CPU, which has 18 cores, and either 2 or 4 Nvidia Tesla K80 GPUs.
We allow $50$ GB of RAM for all models and settings.

\subsection{\label{sec:baselines}Baselines}
\begin{figure*}[htb]
  \centering
  \includegraphics[width=1.0\textwidth]{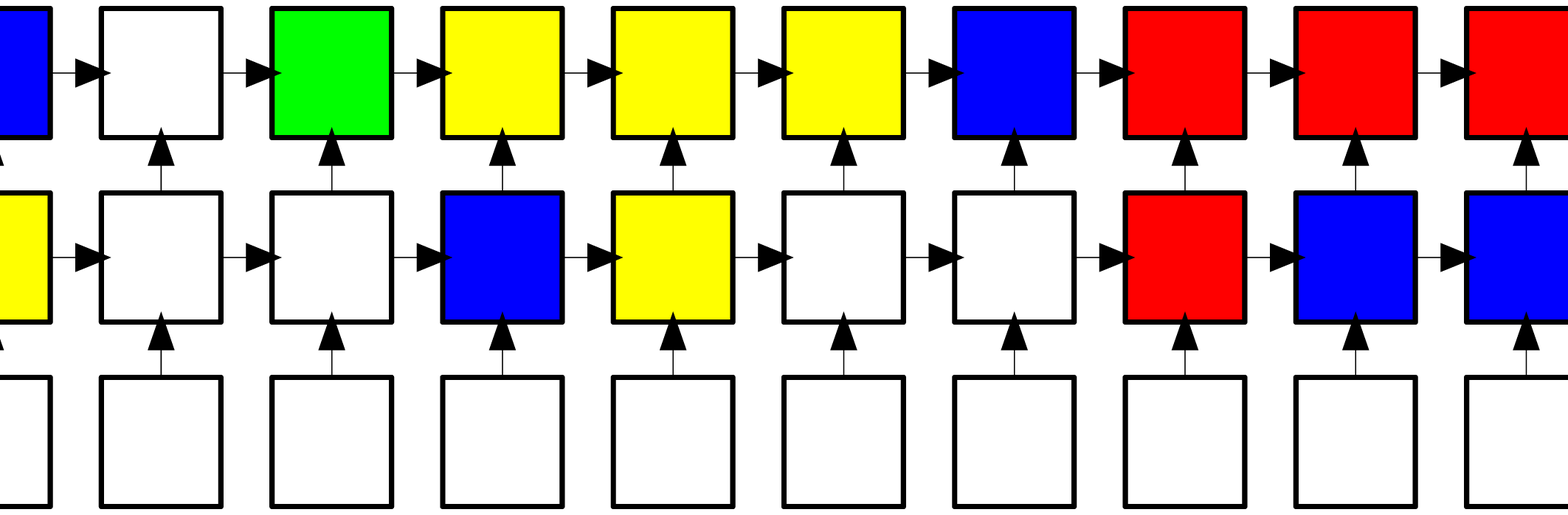}
  \includegraphics[width=1.0\textwidth]{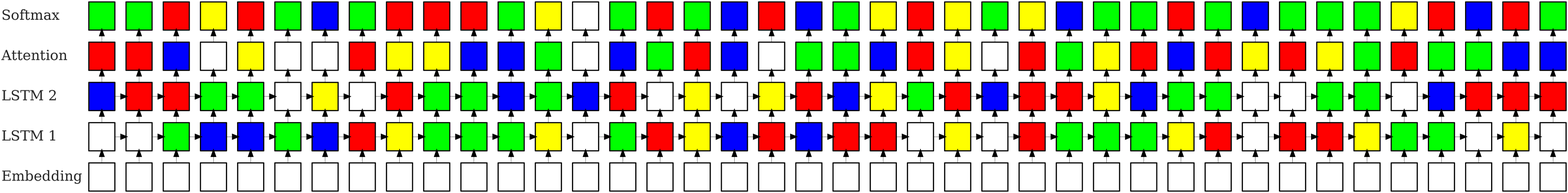}
  \caption{\label{fig:nmt_placement}RL-based placement of Neural MT graph.
  Top: encoder, Bottom: decoder. Devices are denoted by colors, where the
  transparent color represents an operation on a CPU and each other unique
  color represents a different GPU. This placement achieves an improvement of
  $19.3\%$ in running time compared to the fine-tuned expert-designed placement.}
\end{figure*}

\paragraph{Single-CPU.} This placement executes the whole neural network on a
single CPU. Processing some large models on GPUs is infeasible due to memory
limits, leaving Single-CPU the only choice despite being slow.

\paragraph{Single-GPU.} This placement executes the whole neural network on a
single CPU. If an operation lacks GPU implemention, it will be placed on CPU.

\paragraph{Scotch.} We estimate the computational costs of each operation as
well as the amount of data that flows along each edge of the neural network
model, and feed them to the Scotch static mapper~\citep{scotch}. We also
annotate the architecture graph (see Section~\ref{sec:related}) with compute
and communication capacities of the underlying devices.

\paragraph{MinCut.} We use the same Scotch optimizer, but eliminate the CPU
from the list of available devices fed to the optimizer.  Similar to the
single-GPU placement, if an operation has no GPU implementation, it runs on the
CPU.

\paragraph{Expert-designed.} For RNNLM and NMT, we put each LSTM layer on a
device. For NMT, we also put the attention mechanism and the softmax layer on
the same device with the highest LSTM layer, and we put the embedding layer on
the same device with the first LSTM layer.  For Inception-V3, the common
practice for the batch size of $32$ is to put the entire model on a single GPU.
There is no implementation of Inception-V3 with batch 32 using more than 1 GPU.
To create an intuitive baseline on multiple GPUs, we heuristically partition
the model into contiguous parts that have roughly the same number of layers. We
compare against this approach in Section~\ref{sec:results}. The common practice
for Inception-V3 with the larger batch size of $128$ is to apply data
parallelism using $4$ GPUs.  Each GPU runs a replica of the model and processes
a batch of size $32$~\citep{inception}. We compare against this approach in
Section~\ref{sec:end_to_end}.

\subsection{\label{sec:results}Single-Step Runtime Efficiency}
In Table~\ref{tab:results}, we present the per-step running times of the
placements found by our method and by the baselines. We observe that our model
is either on par with or better than other methods of placements.  Despite
being given no information other than the running times of the placements and
the number of available devices, our model learns subtle tradeoffs between
performance gain by parallelism and the costs induced by inter-device
communications.

\paragraph{RNNLM.} Our method detects that it is possible to fit the whole
RNNLM graph into one GPU, and decides to do so to save the inter-device
communication latencies. The resulting placement is more than twice faster than
the best published human-designed baseline.

\paragraph{Neural MT.} Our method finds a non-trivial placement (see
Figure~\ref{fig:nmt_placement}) that leads to a speedup of up to $20.6\%$ for
$4$ GPUs. Our method also learns to put the less computational expensive
operations, such as embedding lookups, on the CPU.  We suspect that whilst
being the slowest device, the CPU can handle these lookup operations (which are
less computationally expensive than other operations) to reduce the load for
other GPUs.

\paragraph{Inception-V3.} For Inception-V3 with the batch size of $32$,
RL-based placer learns that when there are only $2$ GPUs available, the degree
of freedom for model parallelism is limited. It thus places all the operations
on a single GPU (although it could use $2$ GPUs). However, when $4$ GPUs are
available, the RL-based placer finds an efficient way to use all of the GPUs,
reducing the model's per-step running time from $4.60$ seconds to $3.85$
seconds. This result is significant, as neither of our baselines could find a
placement better than assigning all the operations to a single GPU.

We also conduct a simple extension of our experiments, by increasing the batch
sizes of RNNLM and NMT to $256$, and their LSTM sizes to $4,096$ and $2,048$,
respectively. This makes the models' memory footprints so large that even one
layer of them cannot be fitted into any single device, hence ruling out the
human-designed placement.  Nevertheless, after several steps of finding
placements that fail to run, our approach manages to find a way to successfully
place input models on devices The running times of the placements found for
large RNNLM and NMT are $33.46$ and $35.84$ seconds, respectively.

\subsection{\label{sec:end_to_end}End-to-End Runtime Efficiency}
We now investigate whether the RL-based placements can speedup not only the
single-step running time but also the entire training process.

\begin{figure*}[htb]
  \centering
  \includegraphics[width=1.0\textwidth]{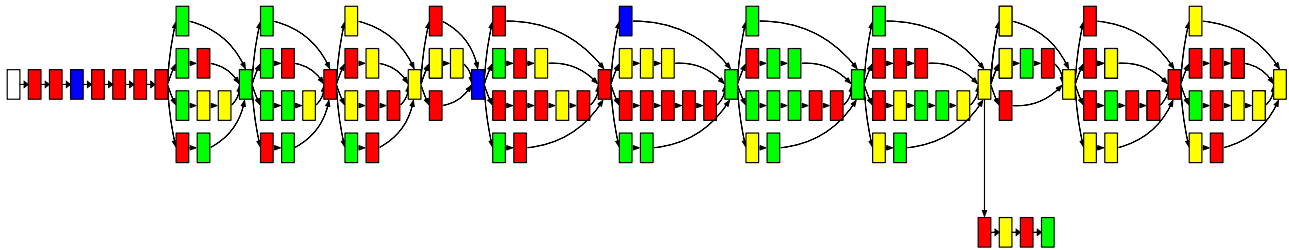}
  \caption{\label{fig:inception_placement}RL-based placement of Inception-V3.
  Devices are denoted by colors, where the transparent color represents an
  operation on a CPU and each other unique color represents a different GPU.
  RL-based placement achieves the improvement of $19.7\%$ in running time
  compared to expert-designed placement.}
\end{figure*}

\paragraph{Neural MT.} We train our Neural MT model on the WMT14 English-German
dataset.\footnote{\url{http://www.statmt.org/wmt14/}} For these experiments, we
pre-process the dataset into word pieces~\cite{gnmt} such that the vocabularies
of both languages consist of $32,000$ word pieces. In order to match our
model's settings, we consider only the translation pairs where no sentence has
more than $40$ word pieces. We train each model for $200,000$ steps and record
their train perplexities. Each training machine has $4$ Nvidia Tesla K80 GPUs
and $1$ Intel Haswell 2300 CPU. Since there are inevitable noises in the
computer systems when measuring the running times, we train each model $4$
times independently and average their per-step running times and perplexities.

\begin{figure}[h!]
  \centering
  \includegraphics[width=0.35\textwidth]{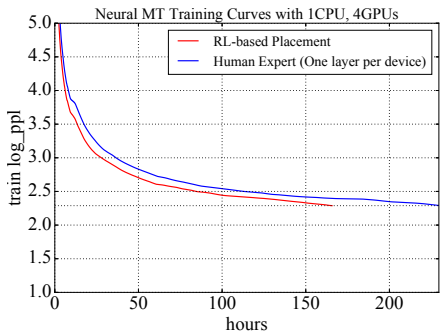}
  \caption{\label{fig:nmt_benchmark}Training curves of NMT model using RL-based
  placement and expert-designed placement. The per-step running time as well as
  the perplexities are averaged over $4$ runs.}
\end{figure}

The RL-based placement runs faster than the expert-designed placement, as shown
in the training curves in Figure~\ref{fig:nmt_benchmark}. Quantitatively, the
expert-designed placement, which puts each layer (LSTM, attention and softmax)
on a different GPU, takes $229.57$ hours; meanwhile the RL-based placement (see
Figure~\ref{fig:nmt_placement}) takes $165.73$ hours, giving $27.8\%$ speed
up of total training time.  We note that the measured speedup rate (and the
running times) of these models appear different than reported in
Table~\ref{tab:results} because measuring them in our RL method has several
overheads.

\paragraph{Inception-V3.} We train Inception-V3 on the ImageNet
dataset~\citep{imagenet} until the model reaches the accuracy of $72\%$ on the
validation set. In practice, more often, inception models are trained with data
parallelism rather than model parallelism. We thus compare the placements found
by our algorithm (see Figure~\ref{fig:inception_placement}) against two such
baselines.

The first baseline, called Asynchronous towers, puts one replica of the
Inception-V3 network on each GPU.  These replicas share the data reading
operations, which are assigned to the CPU. Each replica independently performs
forward and backward passes to compute the model's gradients with respect to a
minibatch of $32$ images and then updates the parameters asynchronously.  The
second baseline, called Synchronous Tower, is the same as Asynchronous towers,
except that it waits for the gradients of all copies before making an update.
All settings use the learning rate of $0.045$ and are trained using RMSProp.
\begin{figure}[h!]
  \centering
  \includegraphics[width=0.35\textwidth]{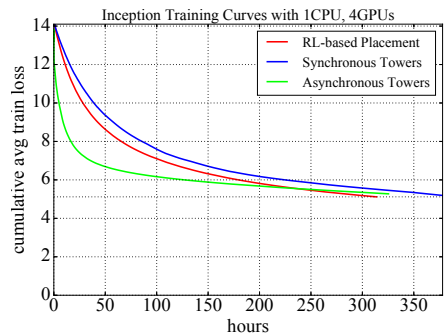}
  \caption{\label{fig:inception_benchmark}Training curves of Inception-V3 model
  using RL-based placement and two expert-designed placements: Synchronous
  towers and Asynchronous towers. The per-step running time as well as the
  perplexities are averaged over $4$ runs.}
\end{figure}

Figure~\ref{fig:inception_benchmark} shows the training curves of the three
settings for Inception-V3. As can be seen from the figure, the end-to-end
training result confirms that the RL-based placement indeed speedups the
training process by $19.7\%$ compared to the Synchronous Tower. While
Asynchronous towers gives a better per-step time, synchronous approaches lead to
faster convergence. The training curve of the RL-based placement, being slower
at first, eventually crosses the training curve of Asynchronous towers.

\subsection{Analysis of Found Placements}
In order to understand the rationale behind the RL-based placements, we analyze
their profiling information and compare them against those of expert-designed
placements.
\begin{figure}[h]
  \centering
  \includegraphics[width=0.40\textwidth]{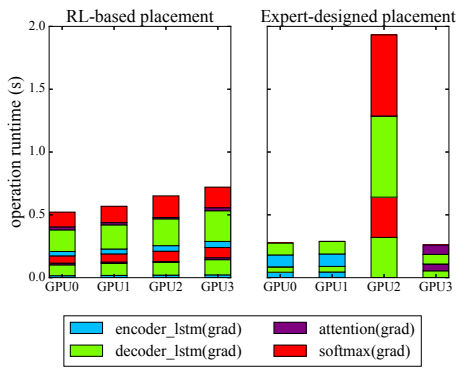}
  \caption{\label{fig:nmt_profiling}Computational load profiling of NMT model
  for RL-based and expert-designed placements.  Smaller blocks of each color
  correspond to feedforward path and same-color upper blocks correspond to
  backpropagation. RL-based placement performs a more balanced computational
  load assignment than the expert-designed placement. }
\end{figure}

\paragraph{Neural MT.} We first compare the per-device computational loads by
RL-based placement and expert-designed placement for the NMT model.
Figure~\ref{fig:nmt_profiling} shows such performance profiling. RL-based
placement balances the workload significantly better than does the
expert-designed placement. Interestingly, if we do not take into account the
time for back-propagation, then expert-designed placement makes sense because
the workload is more balanced (whilst still less balanced than ours). The
imbalance is much more significant when back-propagation time is considered.
\begin{figure}[h!]
    \centering
    \includegraphics[width=0.40\textwidth]{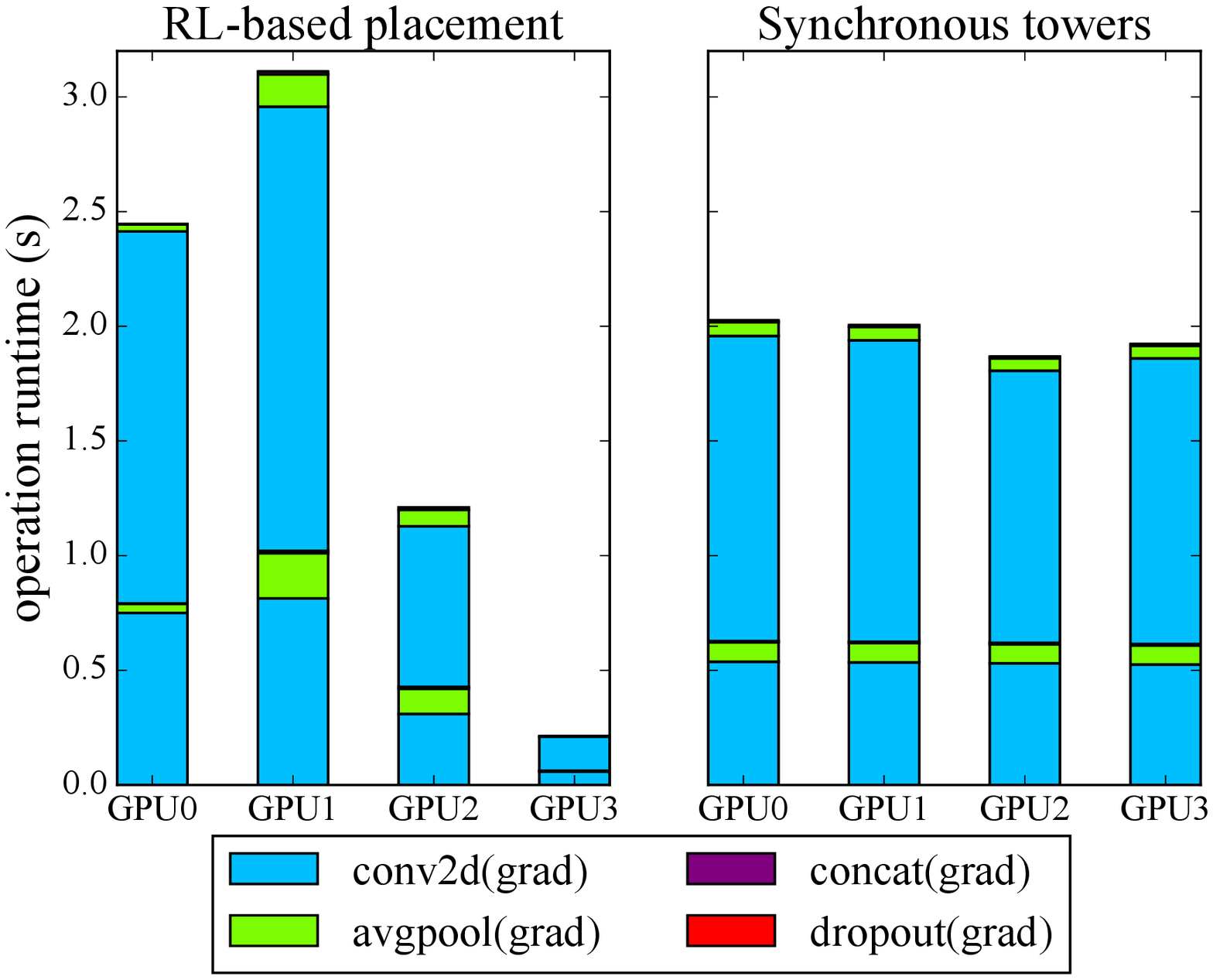}
    \includegraphics[width=0.40\textwidth]{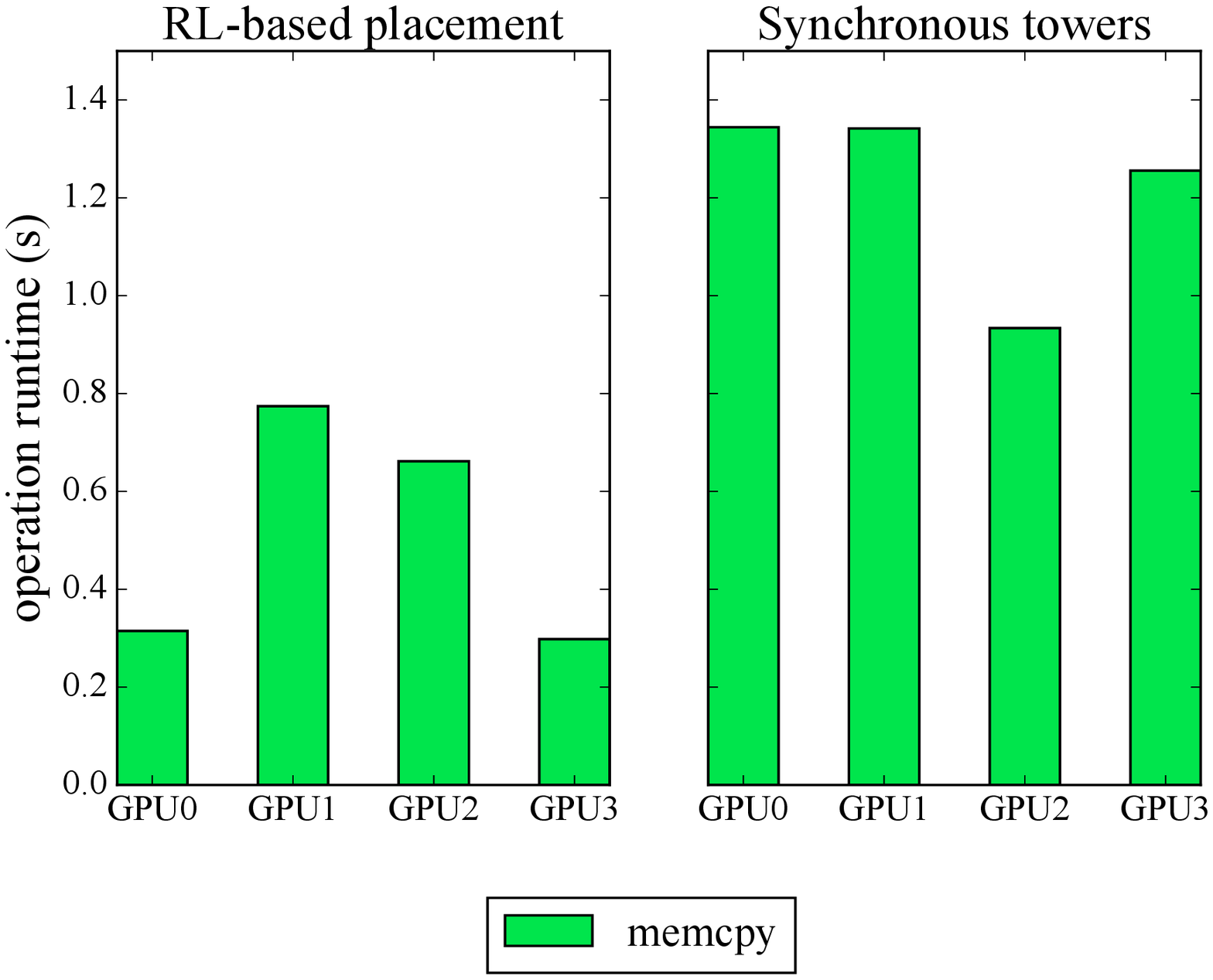}
    \caption{\label{fig:inception_profiling1} Computational load and memory
    copy profiling of Inception-V3 for RL-based and Synchronous tower
    placements. Top figure: Operation runtime for GPUs. Smaller blocks of each
    color correspond to feedforward path and same-color upper blocks correspond
    to backpropagation. RL-based placement produces less balanced computational
    load than Synchronous tower.  Bottom figure: Memory copy time. All memory
    copy activities in Synchronous tower are between a GPU and a CPU, which
    are in general slower than GPU to GPU copies that take place in the
    RL-based placement.}
\end{figure}

\paragraph{Inception-V3.} On Inception-V3, however, the RL-based placement does
not seek to balance the computations between GPUs, as illustrated in
Figure~\ref{fig:inception_profiling1}-top. We suspect this is because
Inception-V3 has more dependencies than NMT, allowing less room for model
parallelism across GPUs.  The reduction in running time of the RL-based
placement comes from the less time it spends copying data between devices, as
shown in Figure~\ref{fig:inception_profiling1}-bottom. In particular, the
model’s parameters are on the same device as the operations that use them,
unlike in Synchronous tower, where all towers have to wait for all parameters
have to be updated and sent to them. On the contrary, that use them to reduce
the communication cost, leading to overall reduction in computing time.

\section{\label{sec:conclusion}Conclusion}
In this paper, we present an adaptive method to optimize device placements for
neural networks. Key to our approach is the use of a sequence-to-sequence model
to propose device placements given the operations in a neural network. The
model is trained to optimize the execution time of the neural network.  Besides
the execution time, the number of available devices is the only other
information about the hardware configuration that we feed to our model.

Our results demonstrate that the proposed approach learns the properties of the
environment including the complex tradeoff between computation and
communication in hardware. On a range of tasks including image classification,
language modeling, and machine translation, our method surpasses placements
carefully designed by human experts and highly optimized algorithmic solvers.

\section*{Acknowledgements}
We thank Martin Abadi, Stephan Gouws, and the Google Brain team for their help
with the project.

\bibliography{main}
\bibliographystyle{icml2017}

\end{document}